\UseRawInputEncoding
\documentclass[verbose=true,letterpaper,margin=1in]{article}

\PassOptionsToPackage{numbers, compress}{natbib}
\usepackage[preprint]{neurips_2024}

\usepackage[utf8]{inputenc} % allow utf-8 input
\usepackage[T1]{fontenc}    % use 8-bit T1 fonts
\usepackage{hyperref}       % hyperlinks
\usepackage{url}            % simple URL typesetting
\usepackage[table]{xcolor} % color for tables
\usepackage{amsfonts}       % blackboard math symbols
\usepackage{nicefrac}       % compact symbols for 1/2, etc.
\usepackage{microtype}      % microtypography
\usepackage{xcolor}         % colors

\usepackage{caption}
\usepackage{amsmath}
\usepackage{multirow}
\usepackage{arydshln}
\usepackage{makecell}
\usepackage{colortbl}
\usepackage[table]{xcolor}
\usepackage{enumitem}
\usepackage{subcaption}

\usepackage[T1]{fontenc}    % use 8-bit T1 fonts
\usepackage{hyperref}       % hyperlinks
\usepackage{url}            % simple URL typesetting
\usepackage{booktabs}       % professional-quality tables
\usepackage{graphicx,amsmath,amssymb} 

\usepackage[ruled,vlined,linesnumbered]{algorithm2e}

\usepackage[utf8]{inputenc}
\usepackage{geometry}           % 调页边距可选
\usepackage{tcolorbox}          % 彩色框
\tcbuselibrary{listings,breakable} % tcolorbox 子库：支持 listings 与自动换行
\usepackage{listings}           % 代码/纯文本高亮
\newtcblisting{promptbox}[2][]{      % #2 = 标题
  listing engine=listings,
  title={#2},
  coltitle=white,
  colbacktitle=black,
  colback=white,
  colframe=black,
  fonttitle=\bfseries,
  arc=3mm,              % 圆角
  outer arc=3mm,
  boxrule=0.8pt,
  breakable,            % 允许跨页
  listing only,         % 内容按 verbatim 处理
  listing options={basicstyle=\ttfamily\small,breaklines=true},
  #1
}

\definecolor{goldenorange}{HTML}{D79F2F}

\title{Beyond `Aha!': Toward Systematic Meta-Abilities Alignment in Large Reasoning Models}

\author{
  Zhiyuan Hu$^{1}$\thanks{Zhiyuan Hu. \href{mailto:zhiyuan_hu@u.nus.edu}{Email: zhiyuan\_hu@u.nus.edu}} \quad 
  Yibo Wang$^{2}$ \quad 
  Hanze Dong$^{3}$ \quad 
  Yuhui Xu$^{3}$ \\
  \textbf{Amrita Saha}$^{3}$ \quad 
  \textbf{Caiming Xiong}$^{3}$ \quad 
  \textbf{Bryan Hooi}$^{1}$\thanks{Corresponding Authors.} \quad 
  \textbf{Junnan Li}$^{3}$\footnotemark[2] \\
  $^1$ National University of Singapore \quad
  $^2$ Tsinghua University \\
  $^3$ Salesforce AI Research
}

\begin{document}

\maketitle

\begin{abstract}

Large reasoning models (LRMs) already possess a latent capacity for long chain-of-thought reasoning. Prior work has shown that outcome-based reinforcement learning (RL) can incidentally elicit advanced reasoning behaviors such as self-correction, backtracking, and verification--phenomena often referred to as the model’s ``aha moment.” However, the timing and consistency of these emergent behaviors remain unpredictable and uncontrollable, limiting the scalability and reliability of LRMs' reasoning capabilities. To address these limitations, we move beyond reliance on prompts and unpredictable `aha moments'. Instead, we explicitly align models with three meta-abilities—\textbf{deduction, induction, and abduction}, using automatically generated, self-verifiable tasks. Our three-stage pipeline (individual alignment, parameter-space merging, domain-specific reinforcement learning) boosts performance by over 10\% relative to instruction-tuned baselines. Furthermore, domain-specific RL from the aligned checkpoint yields an additional gain in performance ceiling for both 7B and 32B models across math, coding, and science benchmarks, demonstrating that explicit meta-ability alignment offers a scalable and dependable foundation for reasoning. Our code is released here. \footnote{\href{https://github.com/zhiyuanhubj/Meta-Ability-Alignment}{https://github.com/zhiyuanhubj/Meta-Ability-Alignment}}

% Drawing on Peirce’s inference triad, we introduce a suite of three lightweight RL environments that individually target deduction, induction and abduction, furnishing both diagnostic signals and training rewards. Each environment produces a specialist model; we then merge their weights with a simple, reproducible recipe to obtain a unified LRM whose complementary strengths raise aggregate accuracy by 20\% points over a standard instruction‑tuned baseline on diverse reasoning benchmarks. Starting further fine‑tuning from this meta‑ability aligned checkpoint, rather than from the vanilla baseline, unlocks an additional 5\% absolute gain, indicating that aligned meta‑abilities form a reusable substrate for downstream capability orchestration. Our results demonstrate that explicit, modular training of fundamental reasoning modes provides a controllable alternative to emergent phenomena, paving the way for systematic skill composition at scale.

\end{abstract}

\section{Introduction}

Large reasoning models, including OpenAI-o1 \citep{jaech2024openai}, o3 \citep{OpenAIo3}, DeepSeek-R1 \cite{guo2025deepseek}, Grok 3.5 \cite{xai2025grok35}, and Gemini 2.5 Pro \cite{google2024gemini25pro}, have demonstrated remarkable capabilities. These models excel at generating long Chain-of-Thought (CoT) \cite{wei2022chain} responses when tackling complex tasks and exhibit advanced, reflection-like reasoning behaviors. Recently, DeepSeek-R1 has shown that, starting from pretrained base or instruction-tuned models, pure reinforcement learning (RL) with rule-based rewards can spontaneously lead to the emergence of long CoT reasoning, self-correction, self-reflection, and other advanced behaviors, collectively referred to as the “aha moment”. Other open-source works, such as SimpleRL-Zoo \citep{zeng2025simplerl}, tinyzero \cite{tinyzero}, and Logic-RL \citep{xie2025logic}, which attempt to reproduce R1's performance and technical details, have also observed similar aha moments. These behaviors—such as self-correction, self-verification, and backtracking, signal the model’s internal experience of strong reasoning ability.

% However, relying solely on such emergent phenomena is inherently unpredictable and uncontrollable. Models may fail to consistently manifest the desired reasoning schemes, which are crucial for reliable problem solving. This unpredictability could hinder the scalability and ceiling of model scaling up and performance. 
% Rather than passively awaiting emergent behaviour, we explicitly \textbf{align the model with three human meta-abilities}, understood as domain-general reasoning \textcolor{red}{modules/element/principle?} that can be recombined across tasks. In this work, we endow large reasoning models with three type of meta-abilities based on Peirce’s classical inference triad \citep{peirce1931collected} - deduction, induction, and abduction.
However, relying solely on emergent behaviors is inherently unreliable and difficult to control. Models may fail to consistently manifest these advanced reasoning schemes, which limits both the predictability and scalability of LLM-based reasoning. To overcome this, we propose to explicitly align LLMs with three domain-general reasoning meta-abilities—deduction, induction, and abduction—drawn from Peirce’s classical inference triad \citep{peirce1931collected}.

% Deduction takes a hypothesis $H$ and a rule set $R$ to derive an observation $O$ $(H + R \rightarrow O)$; induction abstracts the governing rule $R$ from repeated joint occurrences of $H$ and $O$ $(H + O \rightarrow R)$; abduction proposes the most plausible hypothesis $H$ that accounts for a surprising observation $O$ under rules $R$ $(O + R \rightarrow H)$.
% Together this inference triad closes the hypothesis–test–revise loop and equips the model with a controllable pathway through the full scientific reasoning cycle, supporting scalable and reliable problem solving.
% \hanze{These three meta-abilities provide a unified cognitive framework for reasoning:
% Deduction infers specific outcomes from general rules and hypotheses $(H + R \rightarrow O)$, enabling rigorous prediction and verification. Induction abstracts rules from repeated co-occurrences $(H + O \rightarrow R)$, facilitating pattern discovery and generalization.
% Abduction infers the most plausible explanation for surprising observations $(O + R \rightarrow H)$, promoting creative and backward reasoning.
% Together, they form a closed inferential loop for hypothesis generation, testing, and revision, mirroring the scientific method and supporting robust and interpretable reasoning.
% }

\begin{minipage}[t]{0.58\textwidth}
\vspace{1em}

Deduction infers specific outcomes from general rules and hypotheses $(H + R \rightarrow O)$, enabling rigorous prediction and verification. Induction abstracts rules from repeated co-occurrences $(H + O \rightarrow R)$, facilitating pattern discovery and generalization.
Abduction infers the most plausible explanation for surprising observations $(O + R \rightarrow H)$, promoting creative and backward reasoning.

\vspace{1em}

Together, they form a closed inferential loop for hypothesis generation, testing, and revision, mirroring the scientific method and supporting robust and interpretable reasoning.

\end{minipage}%
\hfill
\begin{minipage}[t]{0.35\textwidth}
\vspace{0pt}
\centering
\includegraphics[width=\textwidth]{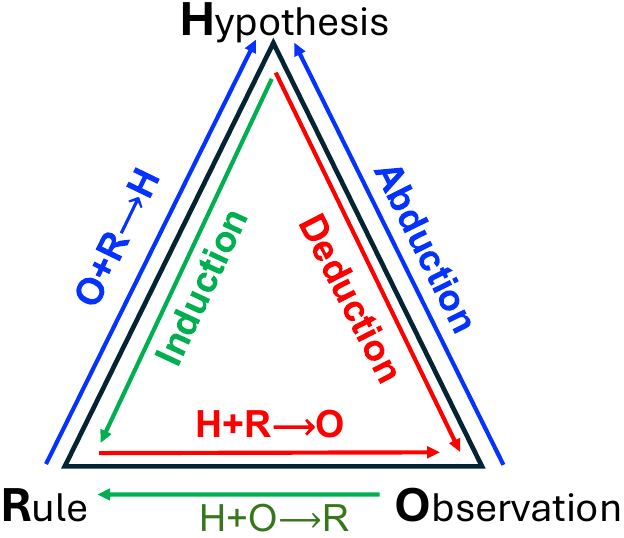}
% \vspace{-3mm}
\captionof{figure}{These meta-abilities form a unified reasoning framework.}
\end{minipage}

% We instantiate three automatically generated, self-verifiable tasks, each honing a distinct meta-ability in Peirce’s inference triad. A single (H, R, O) triple is systematically recast into three “give-two-infer-one” configurations: \textbf{Propositional satisfiability} uses explicit rules R and candidate truth assignments H to test whether all formulae jointly yield the observation O, thereby developing deduction; \textbf{masked-sequence completion} presents visible observations O together with a placeholder for H, so the model must abstract the governing rule R, exercising induction; \textbf{inverse rule-graph search} fixes O and the rule graph R and requires back-chaining to the minimal premise set H, activating abduction. Because every instance is programmatically sampled and accompanied by a ground-truth checker, the suite scales to a large number of examples without human annotation. Moreover, the synthetic distributions are deliberately out-of-distribution relative to web pre-training and standard SFT corpora, preventing leakage of prior heuristics and ensuring that performance gains stem from genuine meta-ability acquisition rather than memorization or domain overlap. \hanze{

To operationalize these meta-abilities, we construct a task suite with programmatically generated instances and automatic verifiability. Each task targets one core reasoning mode:
Deduction: Propositional satisfiability tasks use rule sets  $R$ and candidate hypotheses 
$H$ to test if all premises entail the observation $O$. Induction: Masked-sequence completion requires models to infer latent rules 
$R$ from partial inputs  $H,O$. Abduction: Inverse rule-graph search backchains from observed consequences 
$O$ through a rule graph  $R$ to infer the minimal explanatory  $H$. These tasks are constructed from synthetic distributions that lie out-of-distribution relative to common pretraining corpora, ensuring that performance improvements reflect genuine meta-ability acquisition rather than memorization or shortcut exploitation.

We observe that models aligned to individual meta-abilities make complementary errors. Aggregating their predictions raises overall accuracy by more than 10\% relative to a vanilla instruction-tuned baseline. To incorporate the three competencies into a single network, Rather than training the model on a mixed task corpus, we utilize parameter-space model merging to integrate these meta-abilities. Parameter-space merging improves average accuracy across math, coding, and science by ~2\% on a 7B model and ~4\% on a 32B model over the instruction-tuned baseline, demonstrating the strong generalization of merged meta-abilities.

Furthermore, to evaluate whether meta-ability alignment offers a stronger foundation for subsequent learning, we resumed domain-specific RL training from a checkpoint that had already been aligned and compared it with the same procedure applied to an instruction-tuned model. Starting from the meta-ability checkpoint raises the attainable performance ceiling: after identical continual domain-specific RL training, the model achieves an average gain of about 2\% over its instruction-only counterpart. Our key contributions are as follows:
\begin{itemize}[leftmargin=1.5em]
  \item \textbf{Task suite for meta-abilities.}  
        We introduce a novel RL task suite aligned with three classical meta-abilities—deduction, induction, and abduction—each constructed to train and validate domain-general reasoning skills in large models.
  \item \textbf{Recipe for reasoning mastery.}  
We propose a three-stage recipe: (1) independently align models to each meta-ability; (2) merge them via parameter-space integration; and (3) fine-tune with domain-specific RL. This leads to improved generalization and downstream task accuracy.  

  \item \textbf{Upper-bound boost and scalability.}  
We empirically demonstrate that meta-ability alignment raises the performance ceiling: our 7B and 32B models show consistent gains over instruction-tuned baselines, across math, coding, and science benchmarks.

\begin{figure}
    \centering
    \includegraphics[width=0.9\linewidth]{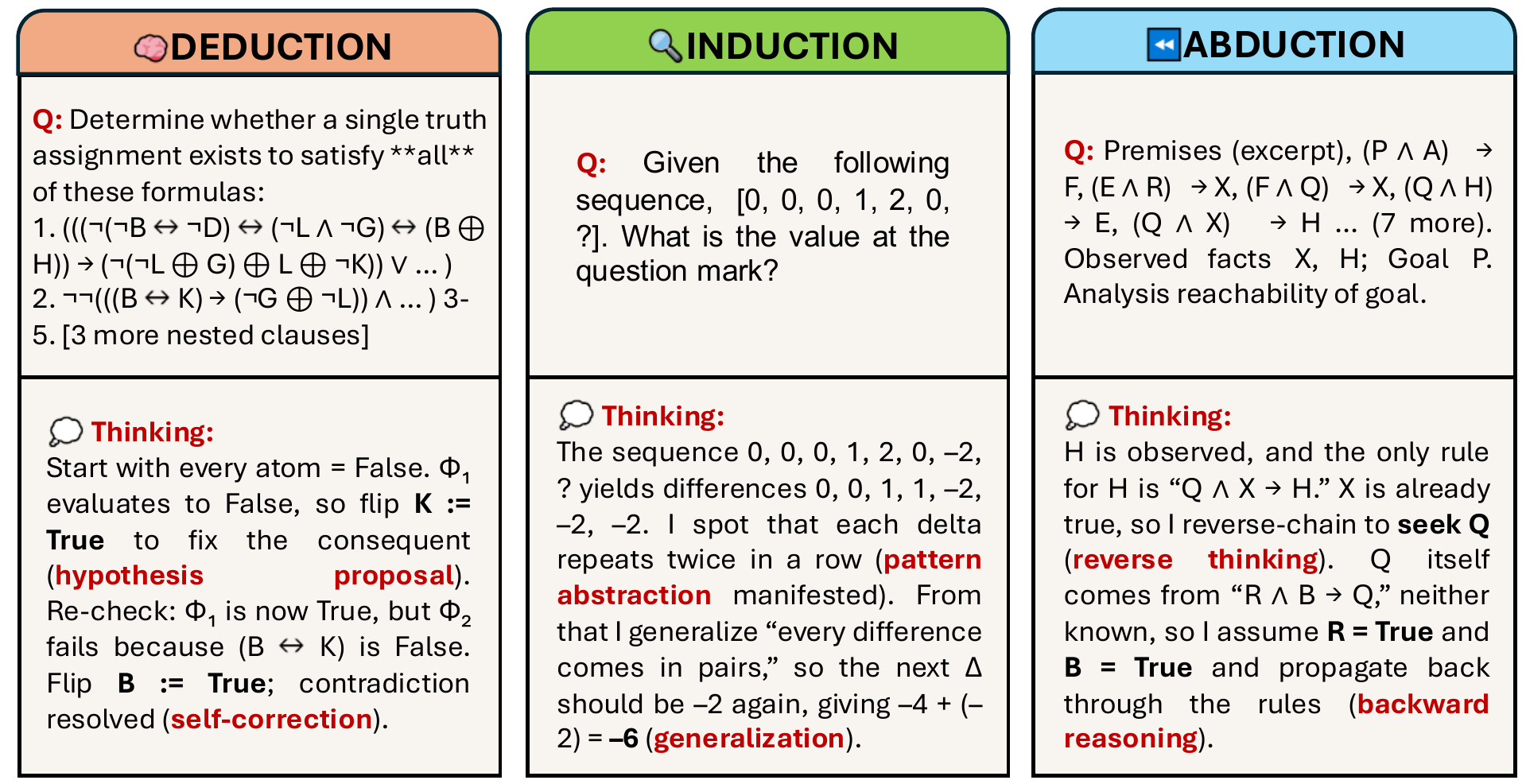}
    \caption{Meta-Ability Alignment Task Example and Corresponding Thought Process}
    \label{fig:enter-example}
\end{figure}

\end{itemize}

\section{Related Work}

\paragraph{RL-Driven Emergence of Reasoning Abilities}  
Recent studies show that direct RL post-training can unlock long chain-of-thought reasoning beyond what supervised fine-tuning achieves \citep{zeng2025simplerl,xiong2025minimalist,guo2025deepseek}.
\emph{SimpleRL-Zoo} \citep{zeng2025simplerl} proposes a zero-RL recipe using rule-based rewards, boosting math reasoning accuracy and inducing cognitive behaviors like self-verification across diverse base models.
\emph{DeepSeek-R1} \citep{guo2025deepseek} extends this idea to large-scale training; its public replications—\emph{Light-R1} \citep{wen2025light}, \emph{Open-R1} \citep{openr1}, and the minimal-cost \emph{TinyZero} \citep{tinyzero} —confirm that curriculum schedules, DPO warm-up, and carefully shaped length rewards together yield stronger logical accuracy while keeping compute affordable.  
Complementary to these general pipelines, \emph{Logic-RL} \citep{xie2025logic} applies rule-conditioned reinforcement learning to synthetic Knights-and-Knaves puzzles, Enabling transferable logical reasoning for math tasks. Together, these works establish RL as a viable path to large reasoning models.

\paragraph{Advanced reasoning ability}  
In addition to enhancing long-chain reasoning through RL, recent work investigates specific reasoning skills such as self-correction, counterfactual inference, self-verification and others.
\citet{chen2024reverse} enhances overall reasoning performance by training models to reason both forward and backward, demonstrating that reverse-thinking objectives can improve forward reasoning as well. \citet{kumar2024training} proposes SCoRe, an online RL method where a model iteratively critiques and improves its own answers, bridging the offline self-correction gap. 
Several recent studies also focus on equipping LLMs with self‑verification and self‑correction abilities, including ProCo \citep{wu2024proco}, S$^{2}$R \citep{ma2025s2r}, and SETS \citep{chen2025sets}.

\paragraph{Investigation of Aha-moment}  
RL pipelines often show sudden accuracy jumps. Some concurrent analyses aim to uncover the internal “aha” moments that precede them. \citet{gandhi2025cognitive} introduces four reasoning behaviors as diagnostic tools to explain and engineer models’ capacity for self-improvement under reinforcement learning. \citet{yang2025understanding} shows that “aha moments” emerge through anthropomorphic language, uncertainty adjustment, and latent-space shifts, helping models avoid reasoning collapse and adapt to problem difficulty. 
Additionally, \citet{zhou2025r1} shows that similar emergence occurs in a 2B vision–language model without supervised warm-up.

\section{Methodology}

\subsection{Task Design for Meta-Abilities Alignment}

\begin{figure}
    \centering
    \includegraphics[width=1\linewidth]{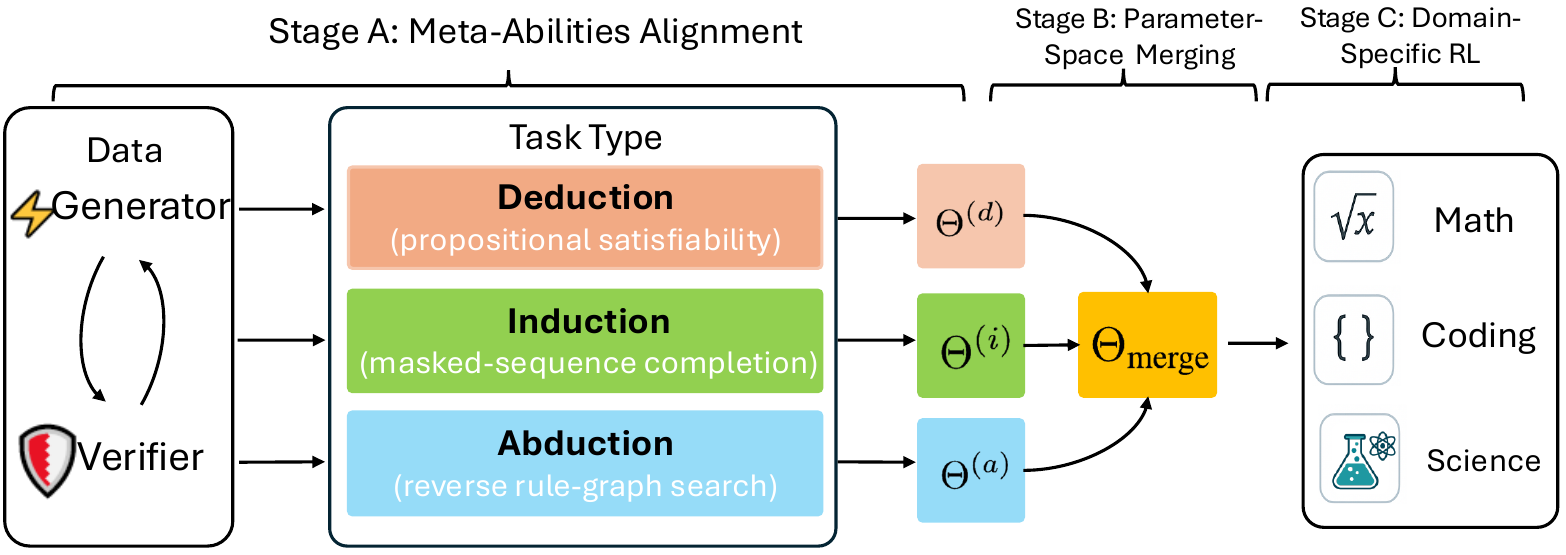}
    \caption{Overview of the three‑stage pipeline: align deduction, induction, and abduction specialists, merge them in parameter space, and continually RL‑adapt the unified model to downstream domains.}
    \label{fig:overview}
\end{figure}

We design three reasoning tasks, each of which involves inferring one element—hypothesis ($H$), rules ($R$), or observation ($O$)—given the other two.
% We design three reasoning tasks by by systematically instantiating the element triad
% $(H, R, O)$ into a “given two, infer the third” framework, each corresponding to a distinct reasoning mode. 
In \textit{deduction} ($H + R \Rightarrow O$), the model is given a set of logical rules $R$ and a candidate truth assignment $H$ as hypothesis, and must verify whether the overall observation $O$ (i.e., all formulas being true) follows—formulated as a \textbf{propositional satisfiability} task. In \textit{induction} ($H + O \Rightarrow R$), the model is provided with observable items $O$ and incomplete inputs $H$ (e.g., masked tokens or implied guesses), and must abstract the underlying generative rule $R$ to correctly complete the sequence—framed as a \textbf{masked-sequence completion} task. In \textit{abduction} ($O + R \Rightarrow H$), the model is given observations $O$ and a rule graph $R$, and must trace backward to recover the minimal set of hidden assumptions $H$ that can logically explain the conclusion—posed as a \textbf{reverse rule-graph search} task. This design follows a strict two-known-one-infer schema, clearly ensuring a clean separation of reasoning types, and reformulates all tasks into a unified (H,R,O) triplet format. This enables consistent, comparable, and complementary training signals, systematically equipping the model with a full range of meta-reasoning capabilities. As illustrated in Figure~\ref{fig:overview}, each instance is produced by an automated \textit{Generator} and screened by a \textit{Verifier}, yielding large‑scale, self‑checked training data entirely free of manual annotation. The pseudo code for data synthesis and additional task examples are provided in Appendix~\ref{Pseudocode}.

% To instil deductive reasoning, the model proceeds from a conjectured hypothesis and specific conditions to derive rigorous predictions, thereby enabling systematic hypothesis proposal, empirical testing, and self-correction. We pose task that present \textcolor{red}{a small bundle of nested propositional formulas built from the standard Boolean operators (\¬,\∧,\∨,\->,\↔,\⊕). The model’s job is to output either a truth assignment that satisfies them all or the verdict UNSAT.} The task poses a combinatorial explosion of possible variable assignments, with tightly coupled logical formulas such that the value of one variable may indirectly constrain or determine the values of many others through chains of logical dependencies. This interdependence renders purely enumerative or heuristically guessed assignments overwhelmingly unlikely to satisfy all constraints. The only tractable approach is to begin with a provisional assignment, treat each formula as a logical premise, systematically derive its consequences, identify any resulting contradictions, and revise the assignment accordingly. This iterative loop—of hypothesis generation, logical consequence propagation, empirical inconsistency detection, and corrective refinement—directly instantiates the core structure of deductive reasoning.

% To instill deductive reasoning, the model proceeds from a conjectured hypothesis and specific conditions to derive rigorous predictions, thereby enabling systematic hypothesis proposal, empirical testing, and self-correction. 

\paragraph{Deduction}

As the example shown in Figure~\ref{fig:enter-example}, we pose task that present a concise cluster of nested propositional clauses involving the standard Boolean operators NOT, AND, OR, IMPLIES, IFF, and XOR; the model must either return a satisfying truth assignment or report that the clauses are unsatisfiable. The task poses a combinatorial explosion of possible variable assignments, with tightly coupled logical formulas such that the value of one variable may indirectly constrain or determine the values of many others through chains of logical dependencies. This interdependence renders purely enumerative or heuristically guessed assignments overwhelmingly unlikely to satisfy all constraints. The only tractable approach is to begin with a provisional assignment, treat each formula as a logical premise, systematically derive its consequences, identify any resulting contradictions, and revise the assignment accordingly. This iterative loop—of hypothesis generation, logical consequence propagation, empirical inconsistency detection, and corrective refinement—directly instantiates the core structure of deductive reasoning.

\paragraph{Induction}
To develop inductive reasoning capabilities in models, we design a task for automatically-generated sequence with hidden terms. Each instance presents a series of elements following an undisclosed pattern (including numeric reasoning, symbolic patterns, and multi-step operation cycles) and requires identification of a missing element. This methodology specifically targets induction as the model must extract the underlying regularity governing the visible sequence and apply it to predict unseen values. Inductive learning through such structured sequences enhances the model's fundamental capabilities in abstraction and generalization, which are essential components for robust reasoning across domains. We also provide the example in Figure~\ref{fig:enter-example} about induction task and ability behavior for better explanation.

\paragraph{Abduction}
To cultivate backward reasoning ability, we introduce a reverse rule-graph search task in which forward inference is deliberately obstructed while backward inference remains efficient. Each instance is formulated as a directed rule graph, with atoms as nodes and implications encoded as hyperedges from premise sets to conclusions. Observed facts activate source nodes, while target hypotheses correspond to sink nodes with unknown truth values. By inflating the branching factor in forward chaining, exhaustive exploration becomes computationally infeasible. In contrast, a backward strategy starts from a goal, hypothesizes minimal supporting premises, and verifies them against known facts. This approach can efficiently isolate relevant subgraphs. The design induces repeated cycles of goal-directed hypothesis formation, verification, and revision, thereby fostering the core mechanism of abductive reasoning.

\subsection{Training Recipe for Reasoning}
Figure~\ref{fig:overview} sketches how we transform the emergent “aha” moment into \emph{controllable, composable} meta-abilities: we first carry out \textbf{Meta-Abilities Alignment}, independently training deduction, induction, and abduction specialists on synthetic diagnostics; we then fuse these specialists through \textbf{Parameter-Space Merging} to obtain a single checkpoint that retains their complementary strengths; finally, \textbf{Domain-Specific Reinforcement Learning Training} further refines the merged model on domain-specific data such as math, coding, and social dialogue.

\subsubsection{Stage A: Meta‑Abilities Alignment}
\label{sec:stageA}
\vspace{0.25em}
We curate three synthetic but diagnostic datasets: \emph{Deduction} (propositional satisfiability),
\emph{Induction} (masked-sequence completion),
and \emph{Abduction} (reverse rule-graph search).
For a policy $\pi_\theta$, we adopt the critic-free \emph{REINFORCE++} loss~\citep{hu2025reinforce++}, along with several improvements proposed in the Logic-RL framework~\citep{xie2025logic}.:
\begin{equation}
\mathcal{J}_{\text{R++}}(\theta)=
\frac{1}{|O|}\!\sum_{i=1}^{|O|}
\Bigl[r_i\,\hat{A}_i - \beta\,D_{\mathrm{KL}}
       \!\bigl(\pi_\theta\|\pi_{\mathrm{ref}}\bigr)\Bigr],\quad
\hat{A}_i=\frac{r_i-\mu_r}{\sigma_r},
\label{eq:reinforcepp}
\end{equation}
where $O$ is the response group, $\pi_{\mathrm{ref}}$ is the frozen
instruction model, $r_i$ the scalar reward, and
$\{\mu_r,\sigma_r\}$ are group statistics. 
% \hanze{If not necessary, minimize the pages occupied by formulas from previous works. GRPO is also similar.}

\vspace{0.25em}
Each reward $r_i$ is computed via a rule-based scheme combining \textbf{Format Reward} and \textbf{Answer Reward}. The Format Reward checks structural compliance using regex-based rules: the model must place reasoning in \texttt{<think>} tags and the final answer in \texttt{<answer>} tags. A correct format yields $+1$, while any deviation gives $-1$. The Answer Reward evaluates correctness relative to the task-specific ground truth: a fully correct answer receives $+2$, and an unparseable or missing answer $-2$. Task-specific criteria guide this evaluation. A deduction (Propositional satisfiability) output is correct only if it satisfies all Boolean formulas; an induction (masked-sequence completion) prediction is valid if the predicted term fits the sequence pattern; and an abduction (inverse rule-graph search) answer is accepted when its premises form the minimal consistent causal path from evidence to target. The total reward is then normalized across the group to produce $\hat{A}_i$.

\subsubsection{Stage B: Parameter-Space Merging for Meta-Ability Integration}
\label{sec:stageB}
To unify the strengths of models specialized in distinct meta-abilities, we adopt \emph{parameter-space merging}, which enables:
(i)~a cost-efficient combination of complementary competencies without additional training, and
(ii)~a high-quality initialization for domain-specific fine-tuning in Stage~C.

We denote the parameters of the deduction-, induction-, and abduction-aligned specialists as $\Theta^{(d)}$, $\Theta^{(i)}$, and $\Theta^{(a)}$, respectively. These models, trained separately on their respective meta-abilities, demonstrate highly complementary behaviors—aggregating their predictions. We construct the merged model $\Theta_{\text{merge}}$ by linearly interpolating the weights of the three specialists:
\begin{equation}
\Theta_{\text{merge}} =
  \lambda_d \Theta^{(d)} +
  \lambda_i \Theta^{(i)} +
  \lambda_a \Theta^{(a)}
\label{eq:model_merge}
\end{equation}

We control the contribution of each specialist model via scalar weights $\lambda_d$, $\lambda_i$, and $\lambda_a$. These coefficients determine the relative influence of each meta-ability in the merged model. Notably, uniform weighting is not assumed—unequal allocation may better reflect the asymmetry in task difficulty or generalization benefit across reasoning modes. Optimal weights are selected empirically based on the performance.
% as described in Section~\ref{sec:exp-setup}.

\subsubsection{Stage C:  Domain-Specific Reinforcement Learning Training}
\label{sec:stageC}

To evaluate whether meta-ability alignment provides a stronger foundation for downstream learning, we apply reinforcement learning to the aligned checkpoints using domain-specific data, specifically math tasks. For a fair and controlled comparison with instruction-tuned baselines, we follow the experimental settings of SimpleRL-Zoo~\citep{zeng2025simplerl}. Specifically, we adopt a rule-based reward function that assigns +1 to correct completions and 0 otherwise, and use the Group Relative Policy Optimization (GRPO) objective~\citep{shao2024deepseekmath} in place of more complex objectives such as REINFORCE++. These choices match SimpleRL-Zoo and help isolate the impact of initialization, ensuring performance gains arise from meta-ability alignment rather than optimization differences.

\begin{equation}
\mathcal{J}_{\text{GRPO}}(\theta)=
\frac{1}{\sum{g}|o_g|}\sum_{i=1}^{G}\sum_{t=1}^{|o_i|}
\min\Bigl[
r_{i,t}\hat{A}i,;
\text{clip}(r{i,t};1{-}\epsilon,1{+}\epsilon)\hat{A}i
\Bigr]
-\beta,D_{\mathrm{KL}}(\pi_\theta|\pi_{\mathrm{ref}}),
\label{eq:grpo}
\end{equation}
where $r_{i,t}$ is the per-token importance weight and $\pi_{\mathrm{ref}}$ is a fixed reference model used to regularize deviations.

\section{Experimental Performance}

\subsection{Experimental Setup}

\paragraph{Dataset}

For each meta-ability task we introduce a specific difficulty-controlling parameter.
% \textcolor{red}{one for propositional satisfiability, another for masked-sequence completion, and a third for inverse rule-graph search} (see Appendix~\ref{} for more details). 
Thus, we generate multiple difficulty levels for every task and adopt the curriculum learning strategy that trains the model level by level from easy to hard (More details about difficulty control are in Appendix~\ref{difficulty}.). With this schedule, the 7B model converges by Level 2, and its reward does not improve further at higher levels, so we restrict training to Levels 1–2. The 32B model occasionally benefits from Level 3 but shows unstable reward curves. Therefore, we also use only the first two levels for it. We sample 200 instances per task per level for the 7B model and 2000 instances per task per level for the 32B model. For further domain-specific RL training, we adopt the same dataset as SimpleRL-Zoo \citep{zeng2025simplerl}.

% \paragraph{Evaluation Benchmarks}

% To validate the generalization of these meta-ability, we select 7 benchmarks from math, coding and science domain. In math, we adopt Math500 \citep{}, AIME (Whole set from 1983 to 2024) \citep{} and Olympic (math subset only) \citep{}. Specially, we test our model in AIME 2024 (AIME24) and AMC 2023 (AMC23) for more latest evaluation to avoid the momery of previous data? and we follow the preview setting in SimoleRL-Zoo to calcuate the Avg@32 with Temperature 1 and Top\_p 0.95. For Coding domain, we use LiveCode Bench (LCB) \citep{}. GPQA \citep{} are for science topic.

\paragraph{Evaluation Setup}

To validate the generalization of these meta-ability, we select 7 benchmarks from math, coding and science domain. In math tasks, we utilize MATH‑500 \citep{hendrycksmath2021}, the full AIME 1983–2024 corpus \citep{aime_1983_2024}, the recent AMC 2023 \citep{maa2023amc} and AIME 2024 \citep{maa2024aime} sets and the Olympiad‑level OmniMath subset \citep{gao2025omnimath} as the evaluation benchmark. LiveCodeBench (LCB) is designed for code generation \citep{jain2024livecodebench}, and GPQA \citep{rein2023gpqa} is aimed for graduate‑level science QA. For most benchmarks, we report pass@1 results using temperature 0.0 and top-p 1.0. While, for AIME 2024 and AMC 2023—which contain fewer problems—we report average accuracy (avg@32), computed over 32 samples per problem using temperature 1.0 and top-p 0.95.

\paragraph{Hyperparameter for Three Stage Training}

We utilize \textsc{Verl} \citep{sheng2024hybridflow} for meta-abilities alignment and continual reinforcement learning, and adopt \textsc{MergeKit} \citep{goddard-etal-2024-arcees} for parameter-space merging to integrate distinct meta-abilities. The optimal weighting coefficients are set to $\lambda_d = 1.0$, $\lambda_i = 0.2$, and $\lambda_a = 0.1$. We provide the comprehensive training setup in Appendix~\ref{setup}.

\begin{table}[h]
\fontsize{7}{8}\selectfont
\centering
\caption{Performance of meta‑ability–aligned models, merged ensembles, and oracle upper bounds on 7 benchmarks at both 7B and 32B parameter scales, illustrating consistent gains from scaling}
\vspace{3mm}
\setlength{\tabcolsep}{5pt}      % ← tighter columns
\renewcommand{\arraystretch}{1.15}
\begin{tabular}{clccccccc|cc}
\toprule
\textbf{Size} & \textbf{Model}
  & \multicolumn{5}{c}{\textbf{Math}}
  & \textbf{Coding}
  & \textbf{Science}
  & \multicolumn{2}{|c}{\textbf{Average}} \\
\cmidrule(lr){3-7}\cmidrule(lr){8-8}\cmidrule(lr){9-9}\cmidrule(lr){10-11}
 &  & \textbf{Math500} & \textbf{AIME}
 & \makecell{\textbf{AIME24}\\\tiny(Avg@32)}
 & \makecell{\textbf{AMC23}\\\tiny(Avg@32)}
 & \textbf{Olympic}
 & \textbf{LCB}
 & \textbf{GPQA}
 & \textbf{Math}
 & \textbf{Overall} \\
\midrule
\multirow{6}{*}{\rotatebox{90}{\textbf{7B}}}
& Qwen2.5-7B-Instruct   & 73.0 & 22.4 & 10.7 & 50.8 & 37.3 & 25.7 & 27.2 & 38.8 & 35.3 \\
& Deduction‑Aligned     & 75.8 & 22.6 & 10.2 & 51.4 & 39.3 & 25.8 & 31.5 & 39.9\textcolor{goldenorange}{(+1.1)} & 36.7\textcolor{goldenorange}{(+1.4)} \\
& Induction‑Aligned     & 75.0 & 22.5 & \textbf{11.8} & 52.3 & 37.5 & \textbf{27.0} & 33.0 & 39.8\textcolor{goldenorange}{(+1.0)} & 37.0\textcolor{goldenorange}{(+1.7)} \\
& Abduction‑Aligned     & 75.2 & 22.7 & 11.4 & 49.1 & 38.4 & 26.8 & 31.9 & 39.4\textcolor{goldenorange}{(+0.8)} & 36.5\textcolor{goldenorange}{(+1.2)} \\
& Merged Model          & \textbf{77.8} & \textbf{22.9} & 11.5 & \textbf{52.3} & \textbf{40.4} & 26.0 & \textbf{33.5} & \textbf{41.0\textcolor{goldenorange}{(+2.2)}} & \textbf{37.8\textcolor{goldenorange}{(+2.5)}} \\
\rowcolor{cyan!10}
\cellcolor{white} & Oracle Ensemble & 85.5 & 32.1 & 18.0 & 67.1 & 46.7 & -- & -- & 49.9\textcolor{goldenorange}{(+11.1)} & -- \\
\midrule
\multirow{6}{*}{\rotatebox{90}{\textbf{32B}}}
& Qwen2.5-32B-Instruct  & 79.8 & 31.2 & 15.3 & 62.7 & 45.6 & 39.5 & 38.0 & 46.9 & 44.6 \\
& Deduction‑Aligned     & 83.8 & \textbf{36.9} & 19.4 & 68.5 & \textbf{47.4} & \textbf{42.1} & \textbf{38.6} & 51.2\textcolor{goldenorange}{(+4.3)} & \textbf{48.1\textcolor{goldenorange}{(+3.5)}} \\
& Induction‑Aligned     & 82.6 & 34.8 & 18.6 & 66.2 & 45.8 & 41.7 & 38.4 & 49.6\textcolor{goldenorange}{(+2.7)} & 46.9\textcolor{goldenorange}{(+2.3)} \\
& Abduction‑Aligned     & 83.8 & 33.3 & 17.5 & 65.9 & 46.2 & 41.1 & 38.6 & 49.3\textcolor{goldenorange}{(+2.4)} & 46.6\textcolor{goldenorange}{(+2.0)} \\
& Merged Model          & \textbf{84.2} & 36.0 & \textbf{19.7} & \textbf{69.5} & 46.9 & 41.8 & 38.4 & \textbf{51.3\textcolor{goldenorange}{(+4.4)}} & \textbf{48.1\textcolor{goldenorange}{(+3.5)}} \\

\rowcolor{cyan!10}
\cellcolor{white}
& Oracle Ensemble & 88.1 & 43.2 & 27.3 & 76.8 & 53.0 & -- & -- & 57.7 \textcolor{goldenorange}{(+10.8)}  & -- \\
\bottomrule
\end{tabular}
\label{tab:aligned}
\end{table}

\vspace{3mm}

\begin{table}[h]
\fontsize{7}{8}\selectfont
\centering
\caption{Comparison of 7B- and 32B-scale baseline instruction models and our domain-specific RL variants across math, code, and science benchmarks. \textit{Domain-RL-Ins} denotes continual domain-specific RL starting from instruction model; \textit{Domain-RL-Meta} applies the same RL schedule but from a meta-ability–merged initialization, yielding a higher attainable performance ceiling.}
\vspace{3mm}
\setlength{\tabcolsep}{5pt}      % tighter columns
\renewcommand{\arraystretch}{1.15}
\begin{tabular}{clccccccc|cc}
\toprule
\textbf{Size} & \textbf{Model}
  & \multicolumn{5}{c}{\textbf{Math}}
  & \textbf{Coding}
  & \textbf{Science}
  & \multicolumn{2}{|c}{\textbf{Average}} \\
\cmidrule(lr){3-7}\cmidrule(lr){8-8}\cmidrule(lr){9-9}\cmidrule(lr){10-11}
 &  & \textbf{Math500} & \textbf{AIME}
 & \makecell{\textbf{AIME24}\\\tiny(Avg@32)}
 & \makecell{\textbf{AMC23}\\\tiny(Avg@32)}
 & \textbf{Olympic}
 & \textbf{LCB}
 & \textbf{GPQA}
 & \textbf{Math}
 & \textbf{Overall} \\
\midrule
\multirow{3}{*}{\rotatebox{90}{\textbf{7B}}}
& Qwen2.5-7B-Instruct & 73.0 & 22.4 & 10.7 & 50.8 & 37.3 & \textbf{25.7} & 27.2 & 38.8 & 35.3 \\
& Domain-RL-Ins            & 78.2 & 23.6 & 11.9 & 53.2 & 39.3 & 25.1 & 33.0 & 41.2\textcolor{goldenorange}{(+2.4)} & 37.8\textcolor{goldenorange}{(+2.5)} \\
& Domain-RL-Meta    & \textbf{78.8} & \textbf{27.7} & \textbf{12.6} & \textbf{54.7} & \textbf{41.0} & 25.4 & \textbf{33.1} & \textbf{43.0\textcolor{goldenorange}{(+4.2)}} & \textbf{39.0\textcolor{goldenorange}{(+3.7)}} \\
\midrule
\multirow{3}{*}{\rotatebox{90}{\textbf{32B}}}
& Qwen2.5-32B-Instruct & 79.8 & 31.2 & 15.3 & 62.7 & 45.6 & 37.5 & 38.0 & 46.9 & 44.6 \\
& Domain-RL-Ins             & 83.0 & 36.5 & 18.6 & 67.5 & 46.1 & \textbf{41.8} & 38.2
                       & 50.3\textcolor{goldenorange}{(+3.4)}
                       & 47.4\textcolor{goldenorange}{(+2.8)} \\
& Domain-RL-Meta     & \textbf{84.6} & \textbf{38.2} & \textbf{19.8} & \textbf{70.4} & \textbf{48.7} & 41.6 & \textbf{38.6}
                       & \textbf{52.3\textcolor{goldenorange}{(+5.4)}}
                       & \textbf{48.8\textcolor{goldenorange}{(+4.2)}} \\
\bottomrule
\end{tabular}
\label{tab:continual}
\end{table}

\subsection{Out-of-Domain Generalization of Meta-Abilities}

% Starting from the Qwen2.5-7B-Instruct baseline (for example, 22.4 on AIME), each individually aligned meta-ability, including deduction, induction, and abduction, provides consistent but modest improvements. AIME scores increase slightly to 22.6, 22.5, and 22.7 respectively, and AMC23 improves from 50.8 to over 52 in some cases. 
% The Merged model, which integrates all three abilities, further improves performance, reaching 22.9 on AIME and 52.3 on AMC23. This indicates that combining these reasoning abilities produces cumulative benefits. 
% The Or setting, which takes the best prediction among the three specialized models, shows significant potential. For instance, Math500 increases from 73.0 to 85.5 and AIME from 22.4 to 32.1. This suggests that the three abilities are highly complementary and together define a strong upper performance bound.

% For the larger Qwen2.5-32B-Instruct model, aligning meta-abilities has a more substantial impact. Each specialized model for deduction, induction, and abduction outperforms the baseline on most benchmarks. AIME scores improve from 31.2 to 36.9, 34.8, and 33.3 respectively, and AMC23 rises from 62.7 to as high as 68.5. 
% The Merged model consistently achieves the best performance among single models, reaching 36.0 on AIME and 69.5 on AMC23. This confirms the effectiveness of integrating multiple reasoning abilities. 
% Although Or results are not fully reported for this model, the complementarity observed in the 7B setting suggests that even higher upper bounds are achievable at the 32B scale.

Table \ref{tab:aligned} shows that meta-ability alignment, trained solely on synthetic diagnostic tasks, already transfers to seven unseen benchmarks (covering five math evaluations—\textsc{Math500}, \textsc{AIME}, \textsc{AIME24}, \textsc{AMC23}, and \textsc{Olympic}—as well as the LiveCodeBench coding test and the GPQA science QA set). At the 7B scale, the induction-aligned model provides the largest average improvement, lifting the mean score by 1.7\% (from 38.8\% to 39.8\%), whereas the deduction-aligned model yields the largest single math task gain with a 2.8\% increase on \textsc{Math500} (73.0\% -> 75.8\%). Integrating the three meta-abilities in the Merged model further raises the overall average by 2.5\% (to 37.8\%), confirming that the abilities combine constructively—e.g., boosting LiveCodeBench from 25.7\% to 26.0\% and GPQA from 27.2\% to 33.5\%. An Oracle Ensemble that marks a problem correct if any aligned model succeeds boosts the Math-average by 11.1\% (to 49.9\%), underlining the strong complementarity still to be tapped by better fusion methods.

Scaling to the 32B model amplifies the pattern: each aligned model surpasses the Qwen2.5-32B-Instruct baseline, yielding a mean gain of 3.1\% on the Math-overall metric and 2.6\% on the overall average. For reference, the baseline already scores 46.9\% on Math-overall and 44.6\% overall. Deduction alignment contributes a +4.3\% jump on Math-overall (to 51.2\%), with peaks on \textsc{AIME} (+5.7\%) and \textsc{AMC23} (+5.8\%), plus gains of +2.6\% on LiveCodeBench and +0.6\% on GPQA. Induction- and abduction-aligned models follow with respective Math-overall gains of +2.7\% and +2.4\%, and overall average lifts of +2.3\% and +2.0\%. Additionally, the Merged checkpoint improves the overall average by 3.5\%, with a standout 4.4\% gain on the Math-average (46.9\% -> 51.3\%) and strong single-task performance such as +6.8\% on \textsc{AMC23}. A full Oracle Ensemble run shows an additional 10.8\% lift in the math average (to 57.7\%), indicating that the three reasoning modes remain highly complementary even at larger scale.

\subsection{Scalable Gains from Meta-Abilities Alignment}
Table~\ref{tab:continual} quantifies how embedding meta-ability alignment into our domain-specific RL pipeline raises performance across both 7B and 32B model scales. In the 7B setting, starting from the instruction-tuned baseline yields a math score of 38.8 and overall average of 35.3. When we apply domain-specific RL directly to the instruction model (Domain-RL-Ins), math climbs to 41.2 (a +2.4 absolute gain) and overall to 37.8 (+2.5). Crucially, when we instead initialize RL from our meta-ability–merged checkpoint (Domain-RL-Meta), which has already learned systematic deductive, inductive, and abductive routines. the same RL schedule pushes math to 43.0 (+4.2) and overall to 39.0 (+3.7). The benefit is especially pronounced on the compositional AIME subset (+5.3 vs. baseline) and the Olympic subset (+3.7), while code (LCB) and science (GPQA) remain stable or improve slightly, demonstrating that meta-ability alignment disproportionately accelerates gains on the most challenging reasoning benchmarks.

Scaling up to 32B magnifies these trends. The instruction baseline achieves 46.9 in math and 44.6 overall. Domain-specific RL from that starting point raises math to 50.3 (+3.4) and overall to 47.4 (+2.8). But applying identical RL on top of the meta-ability–merged 32B model yields 52.3 in math (+5.4) and 48.8 overall (+4.2). These results correspond to relative improvements of roughly 7\% (Domain-RL-Ins) and 12\% (Domain-RL-Meta) in math, indicating that the meta-aligned model not only begins at a higher performance ceiling but also extracts more value from task-specific fine-tuning. In other words, explicitly teaching core reasoning processes before domain adaptation consistently broadens the upper bound of attainable performance—and this margin widens with model capacity, underscoring the scalable advantage of meta-abilities alignment.

\begin{figure}[htbp]
    \vspace{3mm}
  \centering
  \begin{subfigure}{0.32\textwidth}
    \includegraphics[width=\textwidth]{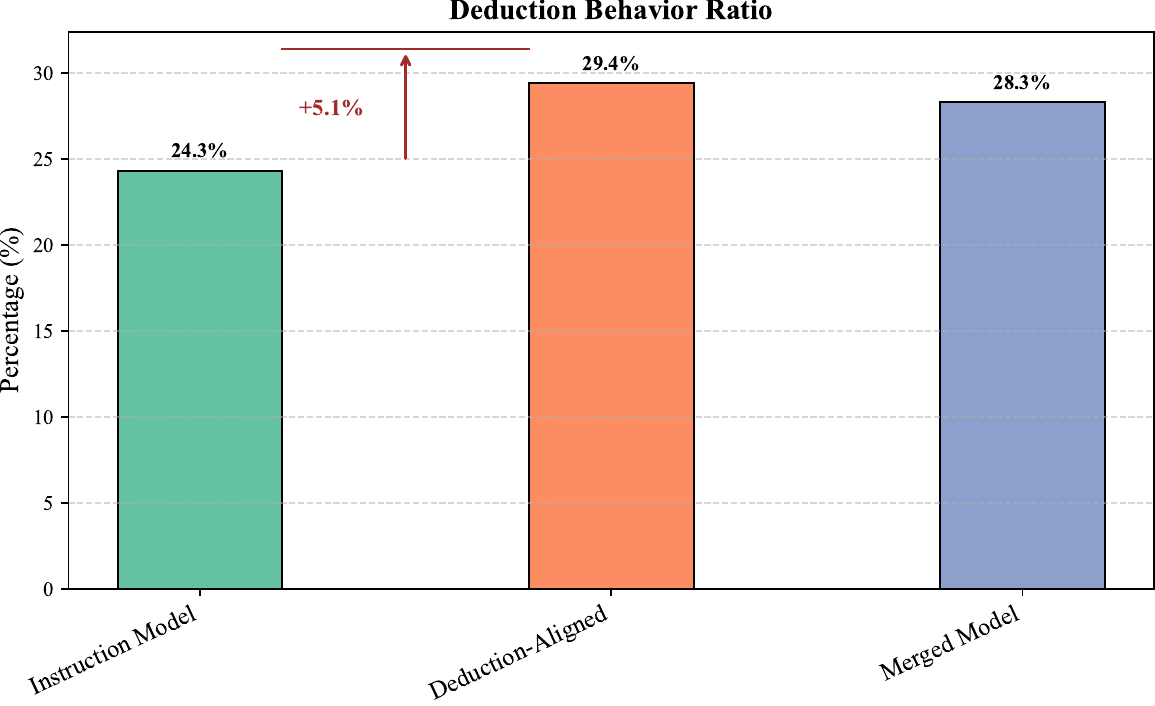}
    % \caption{XXXX}
  \end{subfigure}\hfill
  \begin{subfigure}{0.32\textwidth}
    \includegraphics[width=\textwidth]{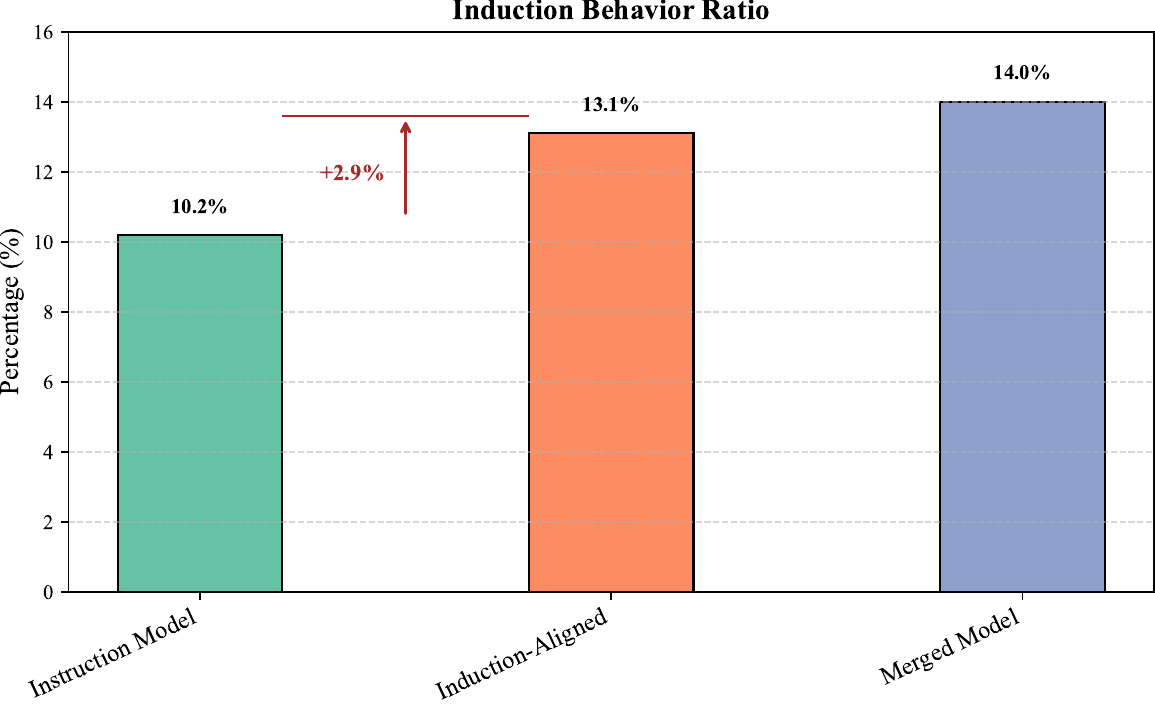}
    % \caption{XXXX}
  \end{subfigure}\hfill
  \begin{subfigure}{0.32\textwidth}
    \includegraphics[width=\textwidth]{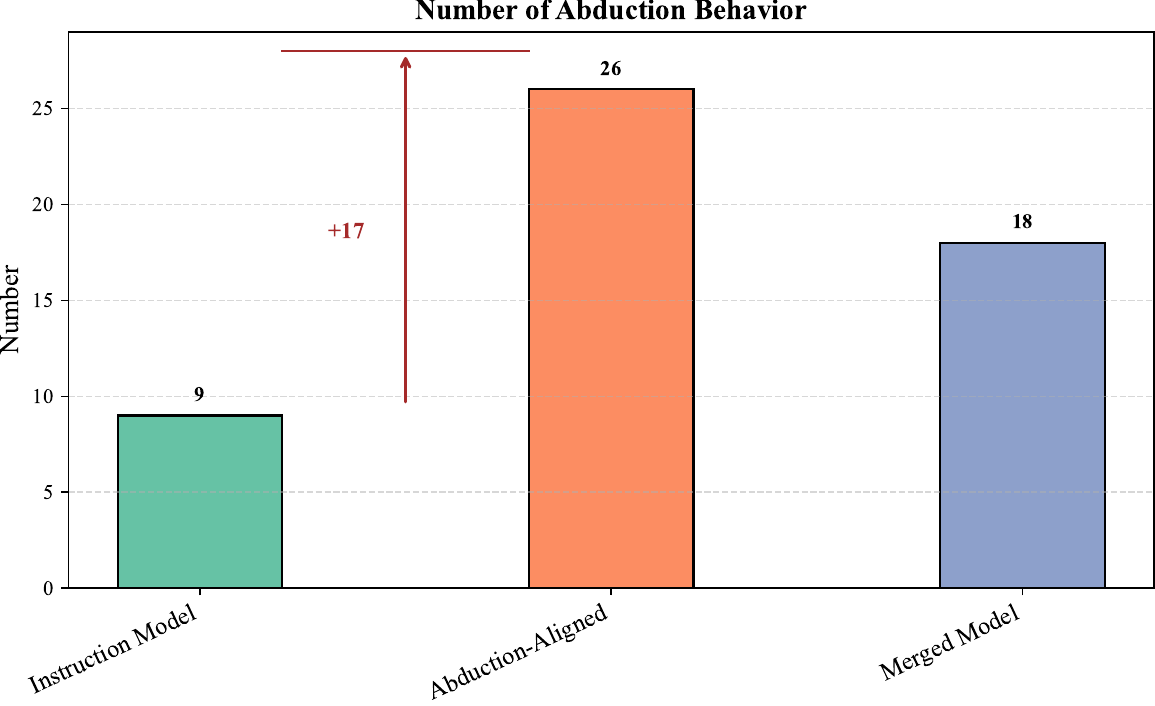}
    % \caption{XXX}
  \end{subfigure}
  \caption{The reasoning behavior ratio(number) for different models}
  \label{fig:three-curves}
\end{figure}

\section{Analysis}

\paragraph{Dose LRM really learn these meta-abilities?}
 To further validate whether meta-ability alignment genuinely enhances the expected advanced reasoning abilities, we adopt the cognitive behavior framework proposed by \citet{gandhi2025cognitive} and utilize OpenAI o4-mini \citep{hurst2024gpt} to identify reasoning-related behaviors. A detailed correspondence between the cognitive behaviors associated with the three meta-abilities, their representative statements and evaluation prompts are provided in Appendix~\ref{prompts}. We report the proportion of model responses that exhibit these cognitive behaviors across the three meta-abilities.

% As shown in Figure \ref{fig:three-curves}, the meta-ability aligned model also exhibits more frequent demonstration of deduction, induction, and abduction than Instruction model.

As Figure~\ref{fig:three-curves} demonstrates, the meta-ability aligned model exhibits a markedly higher frequency of deduction, induction, and abduction compared to the Instruction model, confirming that meta-ability alignment indeed increases the model’s use and expression of these advanced reasoning behaviors.
In particular, deduction occurs most often, which is consistent with our earlier merging strategy that assigned it the highest weight coefficient. Because abduction appears least frequently, we report its raw occurrence count here. The merged model’s performance across all three abilities is at least on par with or slightly better than its pre-merged counterparts.

Second, explicit meta-ability alignment enables us to inject additional desired capabilities into the model—letting us endow it with any specific reasoning skills we choose and tailor its behavior to new application needs.
Third, scientific discovery workflows might profit from models that can reliably hypothesise (abduction), test (deduction), and generalise (induction) in silico before costly lab experiments.

\section{Conclusion}

This work demonstrates that LMRs need not rely on unpredictable “aha moments” to acquire advanced problem-solving skills. By explicitly aligning deduction, induction, and abduction through automatically generated, self-verifiable tasks, we create specialist model whose complementary strengths can be merged—without extra compute—into a single checkpoint that outperforms an instruction baseline by more than 10\% on purpose-built diagnostics and up to 2\% on seven diverse math, code, and science benchmarks. When this meta-ability-aligned model is used as the starting point for domain-specific reinforcement learning, it lifts the attainable performance ceiling by a further 4\% and widens the gap as model capacity scales from 7 B to 32 B parameters; this demonstrates that modular, self-supervised training of core reasoning routines yields gains that generalize reliably beyond narrow task suites. Because each meta-ability is developed via synthetic, verification-driven tasks, no additional human annotation is required—enabling rapid iteration, transparent quality control, and resource-efficient upgrades—while merging specialist checkpoints facilitates flexible composition of reasoning modes tailored to emerging application needs. Isolating deduction, induction, and abduction as discrete modules also supports fine-grained behavior analysis, paving the way for stronger interpretability guarantees and targeted safety audits. By shifting from emergent aha moment to a modular, self-verifying paradigm, we lay the foundation for reasoning systems that are not only more capable but also more predictable, interpretable, and adaptable—charting a path toward safer and more reliable large-scale AI frameworks.

\appendix

\section{Data-Synthesis Pseudocode and Additional Task Examples}
\label{Pseudocode}

\subsection{Deduction}

% ---------- Figure ----------

% ---------- Algorithm ----------
\begin{algorithm}[h]
\caption{\textbf{Propositional Satisfiability}}
\label{alg:dpll}
\DontPrintSemicolon
\KwIn{Clause set $\Phi$ drawn with hyper-parameters 
       $(n_\ell,f_\ell,d_\ell)$ as defined in
       Appendix~\ref{difficulty}.}
\KwOut{A satisfying assignment $\sigma$ or \textsc{UNSAT}.}

\SetKwFunction{DPLL}{DPLL}
\SetKwProg{Fn}{Function}{:}{}

\Fn{\DPLL{$\Phi,\;\sigma$}}{
    \lIf{$\forall C\in\Phi:\;C$ satisfied by $\sigma$}{\Return $\sigma$}
    \lIf{$\exists C\in\Phi:\;C$ falsified by $\sigma$}{\Return \textsc{UNSAT}}
    
    % Unit propagation
    \While{\(\exists\) unit clause \(l\)}{
        $\sigma \gets \sigma \cup \{l\}$;\;
        \(\Phi \gets \textsc{Simplify}(\Phi,l)\)
    }
    
    % Pure-literal elimination
    \ForEach{literal $p$ that appears with only one polarity}{
        $\sigma \gets \sigma \cup \{p\}$;\;
        \(\Phi \gets \textsc{Simplify}(\Phi,p)\)
    }
    
    % Choose branching variable
    pick the first unassigned variable $x$\;
    \For{$b \in \{\textsc{True},\textsc{False}\}$}{
        $\sigma' \gets \sigma \cup \{x{=}b\}$\;
        \(\Phi' \gets \textsc{Simplify}(\Phi,x{=}b)\)\;
        result $\gets$ \DPLL{$\Phi',\sigma'$}\;
        \lIf{result $\neq$ \textsc{UNSAT}}{\Return result}
    }
    \Return \textsc{UNSAT}
}

\BlankLine
\textbf{Main:}\;
\Return \DPLL{$\Phi,\;\emptyset$}\;

\end{algorithm}

\begin{figure}[h]
  \centering
  \includegraphics[width=.8\linewidth]{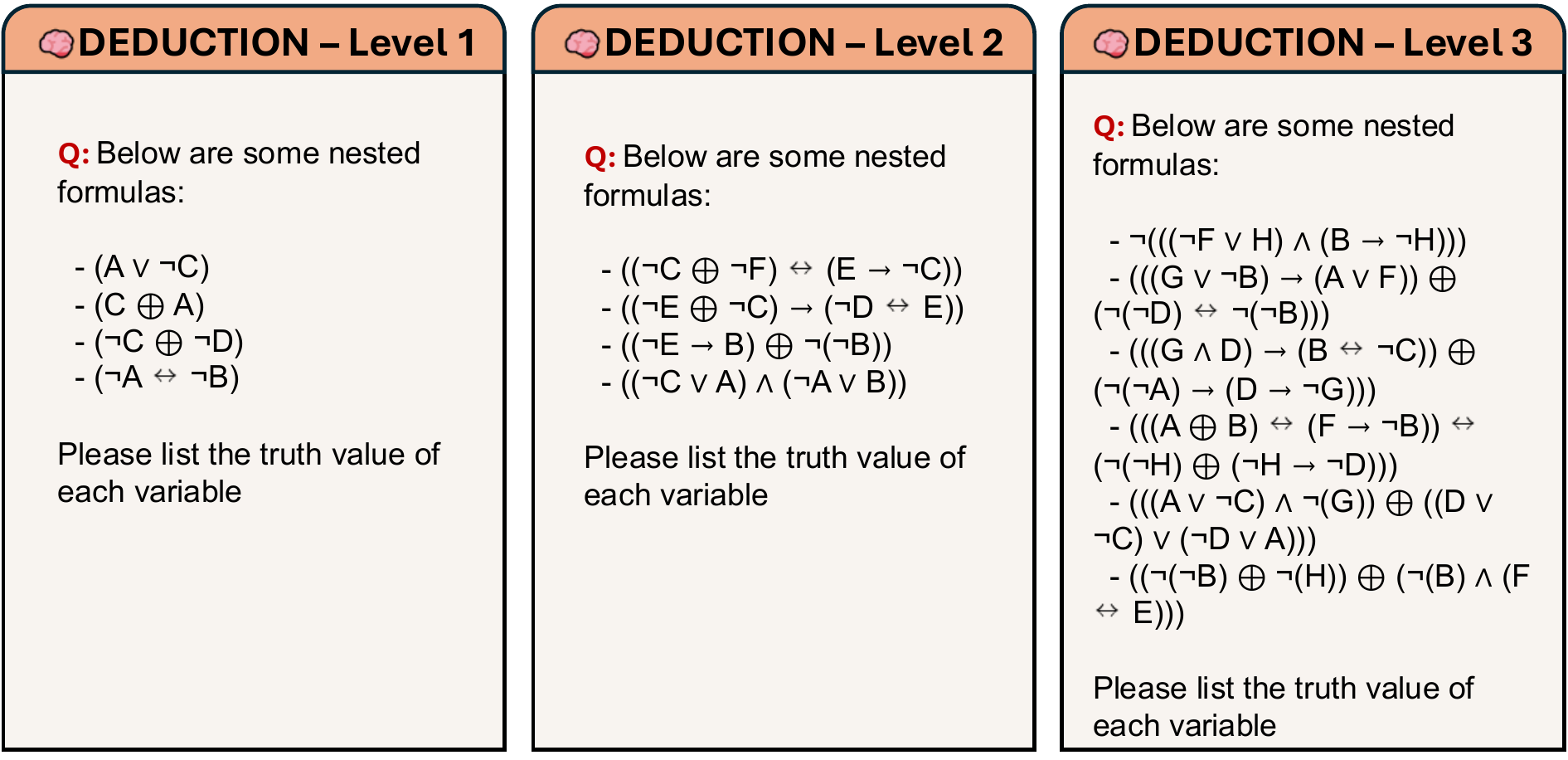}
  \caption{Examples of propositional satisfiability problems with difficulty levels ranging from 1 to 3.}
  \label{fig:deduction-example}
\end{figure}

Figure~\ref{fig:deduction-example} presents examples of propositional satisfiability tasks at difficulty levels 1 through 3, illustrating how clause structures become increasingly complex. While exhaustive truth-assignment trials quickly become intractable, Algorithm~\ref{alg:dpll} begins with unit propagation and pure-literal elimination to assign all forced literals. It then branches on unassigned variables, backtracking when conflicts arise. This process instantiates the hypothesis–consequence–contradiction–refinement cycle characteristic of deductive reasoning, while preserving an exact satisfaction oracle.

\subsection{Induction}

% ---------- Figure ----------
\begin{figure}[b]
  \centering
  \includegraphics[width=.9\linewidth]{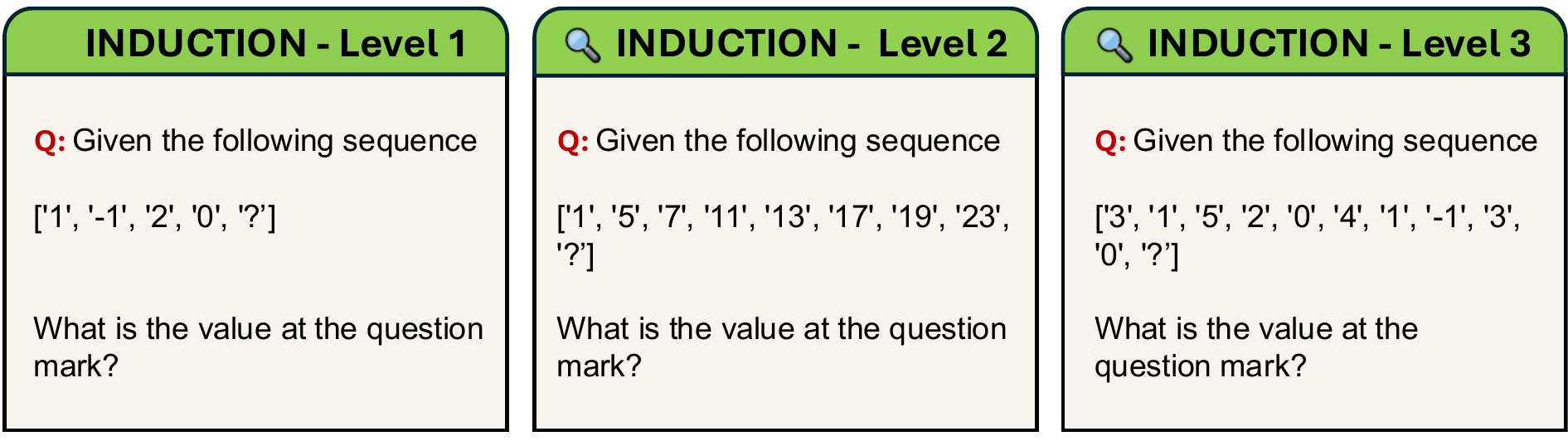}
  \caption{Examples of masked-sequence completion with difficulty levels ranging from 1 to 3.}
  \label{fig:hard-induction}
\end{figure}

% ---------- Algorithm ----------
\begin{algorithm}[H]
\caption{Masked-Sequence Completion}
\label{alg:hard-induce}
\DontPrintSemicolon                      % 不自动显示分号
\SetAlgoLined                            % 显示竖线并自动缩进
\KwIn{Observed prefix $S=[s_1,\dots,s_{n-1},\;?]$; cycle length $k\le 7$;
      operator alphabet $\Sigma\subseteq\{+,-,\times\}\times\{1,\dots,4\}$.}
\KwOut{Predicted missing value}

$\mathcal{B}\leftarrow\{(\langle\rangle,\textsc{score}=0)\}$\;
\For{$t\gets1$ \KwTo $k$}{                        % grow operation sequence
    \If{$\mathcal{B}=\varnothing$}{\Return \textsc{Fail}}   % beam exhausted
    $\mathcal{B}'\leftarrow\varnothing$\;
    \ForEach{candidate $(\vec{o},\textsc{score})\in\mathcal{B}$}{
        \ForEach{$op\in\Sigma$}{
            $\vec{o}'\leftarrow\vec{o}\,\Vert\,op$\;
            \If{\textsc{Consistent}$(\vec{o}',S)$}{
                $\textsc{score}'\leftarrow\textsc{MDL}(\vec{o}')$\;
                add $(\vec{o}',\textsc{score}')$ to $\mathcal{B}'$\;
            }
        }
    }
    keep top $B$ elements of $\mathcal{B}'$ by \textsc{score}\;
    $\mathcal{B}\leftarrow\mathcal{B}'$\;
}
\Return value produced by best $\vec{o}$ when applied to $S$\;
\end{algorithm}

\medskip
Figure~\ref{fig:hard-induction} shows examples of masked-sequence completion tasks at difficulty levels 1 through 3, each featuring a hidden cycle of arithmetic operations of length~$k$ that induces a super‐exponential hypothesis space.
Algorithm~\ref{alg:hard-induce} grows candidate operation sequences up to $k$, prunes those inconsistent with the observed prefix, scores survivors by minimal-description-length, and iteratively refines its beam until the optimal pattern is found, thereby realising the hypothesise–test–refine loop central to inductive reasoning.

\subsection{Abduction}

% ---------- Algorithm ----------
\begin{algorithm}[h]
\caption{\textbf{Reverse Rule-Graph Search}}
\label{alg:backward-search}
\DontPrintSemicolon
\KwIn{Rule-graph $G=(V,E)$; facts $F\subseteq V$; goal $g\in V$;\\
      \textbf{chain\_depth} $d$; \textbf{distractor\_count} $h$; 
      \textbf{cycle\_prob} $\gamma$ (values sampled as in
      Appendix~\ref{difficulty}).}
\KwOut{\textsc{True}/\textsc{False} and (optional) explanation tree $\mathcal{T}$.}

\textbf{Caches:} memo $\mathsf{Cache}$, recursion stack $\mathsf{Path}$,
edge set $\mathcal{E}$.\;

\SetKwFunction{Prove}{Prove}
\SetKwProg{Fn}{Function}{:}{}

\Fn{\Prove{$q,\,\text{depth}$}}{
    \If{$q \in F$}{\Return \textsc{True}}
    \If{\text{depth} $> d$ $\lor\ q\in\mathsf{Path}$}{\Return \textsc{False}}
    \If{$\mathsf{Cache}[q]\neq\textsc{Unknown}$}{\Return $\mathsf{Cache}[q]$}
    
    $\mathsf{Path}\!\leftarrow\!\mathsf{Path}\cup\{q\}$\;
    \ForEach(\tcp*[f]{candidate hyper-edges}){$e:(P\!\rightarrow\!q)\in E$}{
        \If{$\forall p\in P:\,\Prove(p,\text{depth}{+}1)$}{
            $\mathcal{E}\!\leftarrow\!\mathcal{E}\cup\{e\}$\;
            $\mathsf{Cache}[q]\gets\textsc{True}$;
            $\mathsf{Path}\!\leftarrow\mathsf{Path}\setminus\{q\}$;
            \Return \textsc{True}
        }
    }
    $\mathsf{Cache}[q]\gets\textsc{False}$;
    $\mathsf{Path}\!\leftarrow\mathsf{Path}\setminus\{q\}$;
    \Return \textsc{False}
}

\vspace{.3em}
\textbf{Main:}\;
initialise caches;\;
$\textit{success}\gets\Prove(g,0)$;\;
\If{\textit{success}}{prune $\mathcal{E}$ to minimal tree $\mathcal{T}$}
\Else{$\mathcal{T}\gets\varnothing$}
\Return $\textit{success},\,\mathcal{T}$\;

\end{algorithm}

\begin{figure}[t]
  \centering
  \includegraphics[width=.8\linewidth]{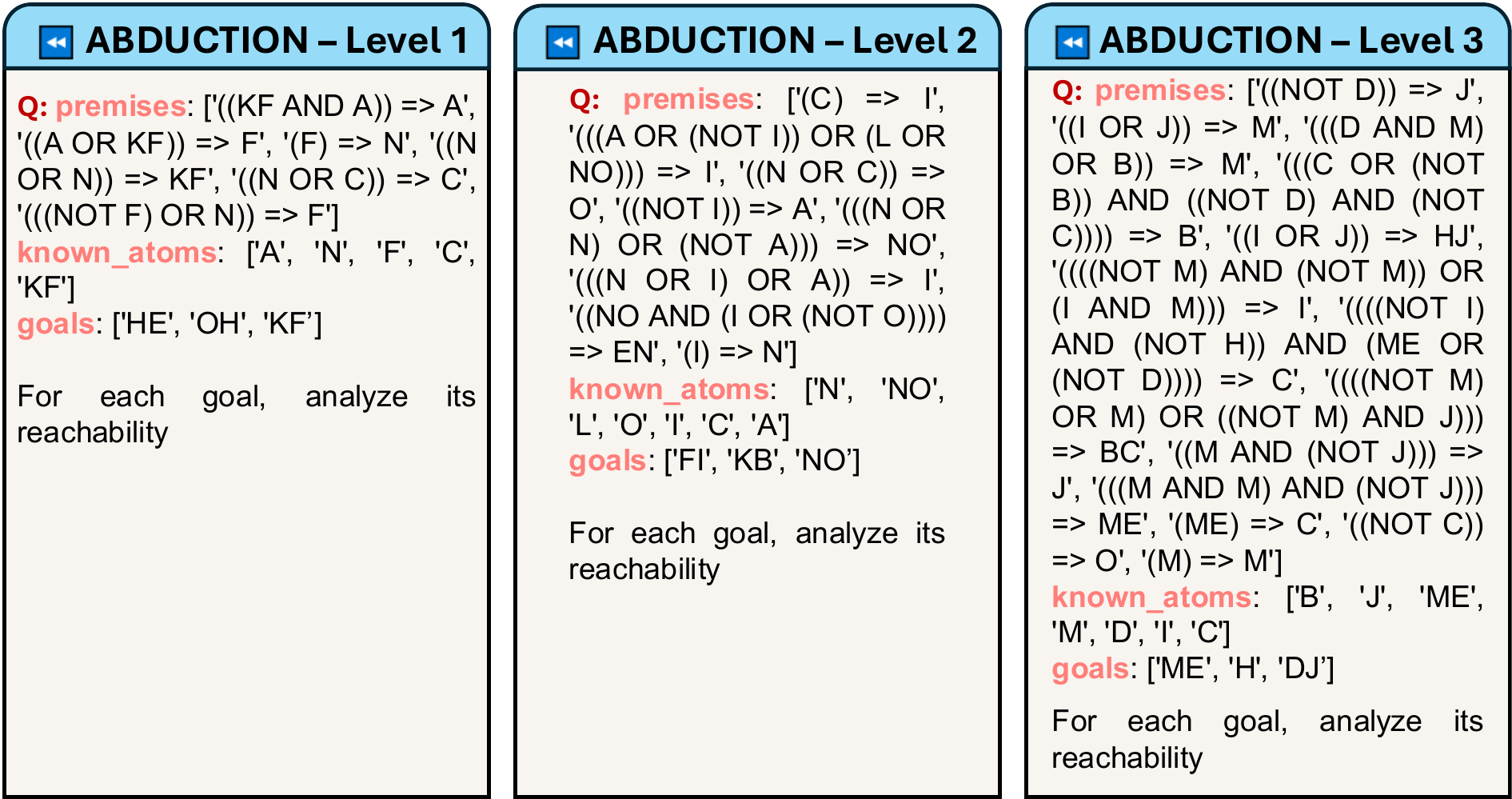}
  \caption{Examples of reverse rule-graph search with difficulty levels ranging from 1 to 3.}
  \label{fig:rrg-example}
\end{figure}

Figure~\ref{fig:rrg-example} shows examples of reverse rule‑graph structures at difficulty levels 1 through 3; Algorithm~\ref{alg:backward-search} begins at the goal \(g\) and recursively traverses candidate edges in reverse, memoizing results and pruning cycles until it either reaches a fact in \(F\) or exceeds the depth limit. The task therefore asks: given facts \(F\) and a goal \(g\), determine reachability and, if successful, return the minimal explanation tree \(\mathcal{T}\) comprising only the utilized hyper‑edges. This goal‑first, premise‑verification loop captures abductive reasoning, while graph depth, branching factor, and distractor rules enable us to scale difficulty yet preserve an exact oracle and dense edge‑level supervision for RL agents.

\section{Controlling Task Difficulty in Synthesis}

\label{difficulty}

We expose \emph{one to four orthogonal hyper-parameters per task} and let the generator sample from progressively wider intervals as the curriculum
advances.  Each hyper-parameter has a clear algorithmic interpretation, so the reader can verify that higher levels provably enlarge the effective search space.

\begin{itemize}

  \item \textbf{Deduction – Propositional Satisfiability.}\par
        Only the nested-formula mode is active in the released code.  A level $\ell$ tuple \(\langle n_\ell,f_\ell,d_\ell\rangle\) controls:
        \vspace{2pt}
        \begin{itemize}
            \item \(n_\ell\): number of propositional variables
                  (\texttt{n\_vars} in the script).
            \item \(f_\ell\): number of independent formulas generated per instance.
            \item \(d_\ell\): maximum parenthesis nesting depth
                  (\texttt{max\_depth}).
        \end{itemize}
        The Boolean assignment space grows as \(2^{n_\ell}\), while deeper nesting couples far-apart variables via implications, making local heuristics ineffective. We empirically observe that moving from $(n,d,f)=(6,2,3)$ to $(10,4,6)$ increases the search time of a basic backtracking solver—which systematically tries possible truth assignments—by two orders of magnitude.

  \item \textbf{Induction – Masked-Sequence Completion.}\par
        For each level we sample \emph{cycle\_length (pattern length, $k$ (1–7))} \(k\in\{1,\dots,7\}\), which is the number of distinct operations that repeat,
        the operator alphabet \(\Sigma\subseteq\{+,-,\times\}\times\{1,\dots,4\}\),
        and the number of masked positions $m$ (1–2).\vspace{2pt}
        \begin{itemize}
            \item A longer cycle $k$ means the model must retain a larger latent state to describe the full rule, e.g.\
                  \(\langle+2,\times2,-3,+4,\times2,-5\rangle\).
            \item Mixing symbolic with numeric operators further enlarges the hypothesis set to \(|\Sigma|^{k}\); at $k\!=\!7$
                  this already exceeds one million templates.
            \item Multiple blanks $m$ break the symmetry that a single missing suffix would have, requiring the agent to interpolate rather than extrapolate.
        \end{itemize}
        Hence the difficulty rises super-exponentially in \(k\) and $m$, emphasizing abstraction and generalization over brute-force
        enumeration.

  \item \textbf{Abduction – Reverse Rule-Graph Search.}\par
        The generator draws
        \texttt{chain\_depth} $d$,
        \texttt{num\_goals} $g$,
        \texttt{distractor\_count} $h$ and
        \texttt{cycle\_prob} $\gamma$ from level-specific ranges, e.g.\
        {\small\texttt{d=3--4}, \texttt{g=2--3}, \texttt{h=5--7},
        \texttt{$\gamma$=0.10–0.25}} for level 3.
        \vspace{2pt}
        \begin{itemize}
            \item Larger $d$ lengthens the \emph{minimal backward proof}, forcing deeper recursion.
            \item Each extra goal $g$ multiplies the number of sink nodes the agent must explain in one episode.
            \item Distractors $h$ are additional hyper-edges whose premises share symbols with real rules; they explode the forward branching factor, so blind forward chaining becomes exponentially slower while a goal-directed agent is unaffected.
            \item A non-zero cycle probability $\gamma$ rewires some edges to form backward loops.  Proof search must therefore keep a visited set and handle \textsc{not-yets}—a hallmark of abductive reasoning.
        \end{itemize}
        Jointly increasing \(\langle d,g,h,\gamma\rangle\) produces a compounded state space whose size we approximate as \(\mathcal{O}((b+\!h)^{d})\) for forward search, where $b$ is the intrinsic out-degree of the knowledge graph.

\item \textbf{Abduction – Reverse Rule-Graph Search.}\par
      The generator draws 
      \texttt{chain\_depth} $d$ (maximal backward-proof depth), 
      \texttt{num\_goals} $g$ (number of sink goals), 
      \texttt{distractor\_count} $h$ (spurious hyper-edges), and 
      \texttt{cycle\_prob} $\gamma$ (edge-rewiring probability)  
      from level-specific ranges, e.g.\ {\small\texttt{d=3–4}, \texttt{g=2–3}, \texttt{h=5–7}, \texttt{$\gamma$=0.10–0.25}} for level 3.
\vspace{2pt}
      \begin{itemize}
        \item Larger $d$ lengthens the \emph{minimal backward proof}, forcing deeper recursion.
        \item Each extra goal $g$ multiplies the number of sink nodes the agent must explain in one episode.
        \item Distractors $h$ share symbols with true rules, exploding the forward branching factor—blind forward chaining slows exponentially, while a goal-directed agent is unaffected.
        \item A non-zero cycle probability $\gamma$ rewires some edges to form backward loops; proof search must therefore maintain a visited set and handle \textsc{not-yets}, a hallmark of abductive reasoning.
      \end{itemize}
      Jointly increasing \(\langle d,g,h,\gamma\rangle\) produces a compounded state space whose size we approximate as \(\mathcal{O}((b+h)^{d})\) for forward search, where $b$ is the intrinsic out-degree of the knowledge graph.

\end{itemize}

\noindent
During curriculum training we feed the model instances in ascending level order (e.g.\ \(d\!=\!2\to7\) for abduction, \(n\!=\!6\to10\) for deduction, \(k\!=\!1\to7\) for induction), so it first acquires stable heuristics on low-entropy regimes before encountering the full combinatorial explosion.

% ---------- Figure ----------

\section{Comprehensive Multi‑Stage Training Setup}

\label{setup}

\paragraph{Stage A: Meta-Abilities Alignment}  
We warm-start from a frozen instruction model and train on three synthetic diagnostic corpora—Deduction (propositional SAT), Induction (masked‐sequence completion) and Abduction (inverse rule‐graph search)—using the critic‐free REINFORCE++ objective.  Rollouts are generated with vLLM (temperature 0.7, $n=4$, tensor‐parallel size 2, max tokens $\approx$ 9 000), and optimized with AdamW (LR $5\times10^{-7}$) over 20 epochs.  We use a training batch size of 32, PPO mini‐batch size of 256, and PPO micro‐batch size of 32.  Rewards are normalized per group to produce $\hat A_i$, and we regularize via a KL penalty ($\beta=0.001$, low‐variance estimator) to the frozen reference model.

\paragraph{Stage C: Domain-Specific Reinforcement Learning on Math}  
Starting from the Stage A checkpoint (after MergeKit parameter merging), we fine-tune on the dataset from SimpleRL Zoo using GRPO objective to match SimpleRL-Zoo settings.  Inputs are capped at 1 024 + 3 072 tokens; train/val batch size 16; PPO mini-batch 128, micro per-GPU 2; rollouts with $n=8$, temperature 1.0, TP size 2.  Rewards are rule-based (+1 for correct, 0 otherwise) without format terms to isolate transfer effects.  We use AdamW (LR $5\times10^{-7}$), clip ratio 0.2, entropy bonus 0.001, and KL to reference ($\beta=0.001$).

\section{Quantifying Meta‑Ability Behaviors through Targeted Prompts}

\label{prompts}

Below we present the three carefully crafted sentence‑extraction prompts—one each for Deduction, Induction and Abduction—that we feed to the model to capture its core reasoning behaviors. Each prompt defines the target micro‑skills, constrains the extraction to only the relevant sentences, and enforces a strict JSON output format so that we can automatically compute the frequency of each reasoning mode. These prompts form the basis for quantifying how often the model engages in each type of advanced inference.
%% ---------------- Deduction ----------------
\begin{promptbox}{Deduction Prompt}
Here is a thought record:

<thinking text>

We define ``Deduction'' to consist of exactly two sub-skills:
  1. HypothesisProposing: sentences that explicitly introduce a testable hypothesis or scenario, such as:
     * ``Assume <\dots>, then <\dots>.''  
     * ``Case <number>: <\dots>.''  
     * ``Suppose <\dots>, then <\dots>.''  

  2. SelfVerification/SelfCorrection: sentences that explicitly validate or correct a prior result, such as:
     * ``After checking, <\dots>, so <\dots>.''  
     * ``We see that <\dots> was wrong, so <\dots>.''  
     * ``This shows an error; we must <\dots>.''  
     * ``Correcting that, <\dots>.''  

Extraction requirements:
- Extract **all and only** the sentences that clearly match one of the above patterns or equivalent wording that directly reflects the two sub-skills.
- Do **not** extract any explanatory, descriptive, or purely procedural sentences.
- Return a JSON array; each element must be exactly:
  {
    "category": "HypothesisProposing"|"SelfVerification/SelfCorrection",
    "sentence": "<the exact extracted sentence>"
  }
- After the array, output one more field:
  "total_count": <the total number of extracted sentences>

Please **strictly** follow this format without any additional commentary.
\end{promptbox}

%% ---------------- Induction ----------------
\begin{promptbox}{Induction Prompt}
You are an expert reasoning analyst.

TASK: Induction Sentence Extraction (Wide-Net, English Only)
----------------------------------------------------------------
1. Input text is wrapped between THINKING: ... END THINKING.
2. Scan every sentence (and its immediately preceding sentence) and extract only those that
   state or imply a general rule, trend, or pattern derived from multiple cases or examples.
   Key English triggers include words or phrases such as:
     * therefore, thus, hence, it follows that
     * we observe that, observations show, we see that
     * typically, usually, in general, most, majority
     * suggests that, implies that, indicates that
     * pattern, trend, rule, conjecture, generalize
     * for all n, for any k, if ... then ...

   Extraction is OK for probabilistic statements (e.g.\ ``Most A are B'') and recursive claims
   (e.g.\ ``If P(k) holds then P(k+1) holds, so P(n) for all n'').
3. Do NOT extract purely numerical computations, single-step deductions, or abduction sentences.

OUTPUT FORMAT (nothing else):
----------------------------------------------------------------
First, a JSON array of the exact extracted sentences as strings. Then, on the next line:
"total_count": <number_of_sentences>

Example:
[
  "For any $n\\equiv 0\\pmod{5}$, the position is losing.",
  "Thus the sequence converges to zero for all starting values."
]
"total_count": 2

THINKING:
<thinking text>
END THINKING

Remember: output valid JSON and the total_count line, and nothing else.
\end{promptbox}

%% ---------------- Abduction ----------------
\begin{promptbox}{Abduction Prompt}
You are an expert reasoning analyst.

**Task**
1. I will give you a block of text labelled
   THINKING: ... END THINKING.
2. Scan every sentence and extract **only** those that exhibit **abduction (backward reasoning)**.
3. Exclude any sentences that use forward reasoning (deduction or induction) or are purely descriptive/mathematical.
4. Return a JSON array; each element must be exactly:
   {
     "sentence": "<the exact extracted sentence>",
     "match_type": "template" | "close_paraphrase" | "clear_abduction_non_template",
     "matched_pattern": "<template code or empty>"
   }
5. After the array, output a final field on a **new line**:
   "total_count": <the total number of extracted sentences>

**Abduction vs.\ Forward Reasoning**
- **Abduction (backward reasoning)** starts from an observation or surprising fact and proposes or tests a hypothesis/explanation.
- **Forward reasoning** (deduction / induction) starts from premises or rules and derives consequences --- **do not extract** those.

**Abduction Templates (code: exact pattern)**
T1 Given-that -> ``Given that <observed fact>, it must be that <hypothesis>.''  
T2 Since-result -> ``Since <result> holds, one explanation is <hypothesis>.''  
T3 To-account-for -> ``To account for <outcome>, suppose <hypothesis>.''  
T4 Observing-suggests -> ``Observing <phenomenon> suggests <hypothesis>.''  
T5 Because-infer -> ``Because <result>, we infer <hypothesis>.''  
T6 Hypothesis-test-revise -> ``Testing that hypothesis, we find <contradiction>, so <revision>.''  
T7 Bad-consequence-fails -> ``However, this would imply <bad consequence>, thus the explanation fails.''  
T8 Closer-inspection -> ``On closer inspection, <hypothesis> does not hold, so <revision>.''  

**Close Paraphrase Examples**
- ``One possible cause of X is Y.''  
- ``A plausible explanation for X could be Y.''  
- ``If X happened, then Y might be responsible.''  
- ``X seems best explained by Y.''  
- ``Assuming Y, we can explain X.''  
- ``Y predicts X; therefore, Y is likely.''  
- ``Y would predict Z, yet we observe not-Z; hence Y is unlikely.''  

**Strict Extraction Rules**
* Extract **all and only** sentences performing abduction as defined above.  
* Exclude any forward-reasoning or purely procedural/mathematical sentences.  
* If a sentence matches a template **exactly**, set `"match_type":"template"` and `"matched_pattern"` to its code (e.g.\ `"T1"`).  
* If it is a tight paraphrase (<= 3 word changes per clause), set `"match_type":"close_paraphrase"`.  
* Otherwise, if it clearly does abduction but is not a template or close paraphrase, set `"match_type":"clear_abduction_non_template"` and leave `"matched_pattern"` blank.  
* Output must be valid JSON with **no** extra keys or explanatory text.

THINKING:
<thinking text>
END THINKING

Please **strictly** follow the specified output format and include nothing else.
\end{promptbox}

\newpage
\bibliographystyle{plainnat}
\bibliography{neurips_2025}

\newpage

\newpage

\end{document}